\documentclass[manuscript]{acmart}

\AtBeginDocument{%
  }

\def\SAdel#1{\bgroup\markoverwith{\textcolor{red}{\rule[0.5ex]{2pt}{1pt}}}\ULon{#1}}

\def\AMdel#1{\bgroup\markoverwith{\textcolor{blue}{\rule[0.5ex]{2pt}{1pt}}}\ULon{#1}}

\def\DZdel#1{\bgroup\markoverwith{\textcolor{green}{\rule[0.5ex]{2pt}{1pt}}}\ULon{#1}}

\usepackage{graphicx}
\usepackage{subcaption}

\setcopyright{acmcopyright}
\copyrightyear{2022}
\acmYear{2022}
\acmDOI{3544999.3552529}

\acmConference[AutoUI'22]{Make sure to enter the correct
  conference title from your rights confirmation emai}{Sept 17--20,
  2022}{Seoul, South Korea}
\begin{document}

\title{To show or not to show: Redacting sensitive text from videos of electronic displays}


\author{Abhishek Mukhopadhyay}
\authornote{This work was carried out during the internship at Pulse Labs AI. \\ Correspondence:  abhishekmukh$@$iisc.ac.in, shubham.agarwal$@$pulselabs.ai, dylan.zwick$@$pulselabs.ai,
pradipta$@$iisc.ac.in}
\affiliation{%
  \institution{CPDM, Indian Institute of Science, Bangalore, Karnataka}
  \streetaddress{Address line 1}
  \city{City}
  \state{State}
  \country{India}}
\email{abhishekmukh@iisc.ac.in}

\author{Shubham Agarwal}
\affiliation{%
  \institution{PulseLabs AI, Salt Lake City, Utah}
  \streetaddress{Address line 3}
  \city{City}
  \state{State}
  \country{United States}}
\email{shubham.agarwal@pulselabs.ai}

\author{Patrick Dylan Zwick}
\affiliation{%
  \institution{PulseLabs AI, Salt Lake City, Utah}
  \streetaddress{Address line 3}
  \city{City}
  \state{State}
  \country{United States}}
\email{dylan.zwick@pulselabs.ai}

\author{Pradipta Biswas}
\affiliation{%
  \institution{CPDM, Indian Institute of Science, Bangalore, Karnataka}
  \streetaddress{Address line 1}
  \city{City}
  \state{State}
  \country{India}}
\email{pradipta@iisc.ac.in}

\renewcommand{\shortauthors}{Mukhopadhyay et al.}
% TODO: SA
% \renewcommand{\shortauthors}{Mukhopadhyay et al.}


\begin{abstract}
With the increasing prevalence of video recordings there is a growing need for tools that can maintain the privacy of those recorded. In this paper, we define an approach for redacting personally identifiable text from videos using a combination of optical character recognition (OCR) and natural language processing (NLP) techniques. We examine the relative performance of this approach when used with different OCR models, specifically Tesseract and the OCR system from Google Cloud Vision (GCV). For the proposed approach the performance of GCV, in both accuracy and speed, is significantly higher than Tesseract. Finally, we explore the advantages and disadvantages of both models in real-world applications.

% In this paper we define an approach for redacting sensitive text information from videos, and compare the performance of two state-of-the-art optical character recognition (OCR) models used in our approach. Our approach is an ensemble of OCR and natural language processing (NLP), and the two OCR models analyzed are Tesseract and the OCR system from Google Cloud Vision (GCV). In our analysis the performance of GCV, in both accuracy and speed, is significantly higher than Tesseract. Finally, we explain the advantages and disadvantages of both models in real-world applications. 

\end{abstract}

\begin{CCSXML}
<ccs2012>
   <concept>
       <concept_id>10010147.10010178.10010179</concept_id>
       <concept_desc>Computing methodologies~Natural language processing</concept_desc>
       <concept_significance>500</concept_significance>
       </concept>
   <concept>
       <concept_id>10010147.10010178.10010179.10003352</concept_id>
       <concept_desc>Computing methodologies~Information extraction</concept_desc>
       <concept_significance>300</concept_significance>
       </concept>
   <concept>
       <concept_id>10010147.10010178.10010224.10010245.10010251</concept_id>
       <concept_desc>Computing methodologies~Object recognition</concept_desc>
       <concept_significance>300</concept_significance>
       </concept>
 </ccs2012>
\end{CCSXML}

\ccsdesc[500]{Computing methodologies~Natural language processing}
\ccsdesc[300]{Computing methodologies~Information extraction}
\ccsdesc[300]{Computing methodologies~Object recognition}

\keywords{Redaction, Optical Character Recognition, Named Entity Recognition, Google Cloud Vision, Tesseract}
\begin{teaserfigure}
  \includegraphics[width=\textwidth]{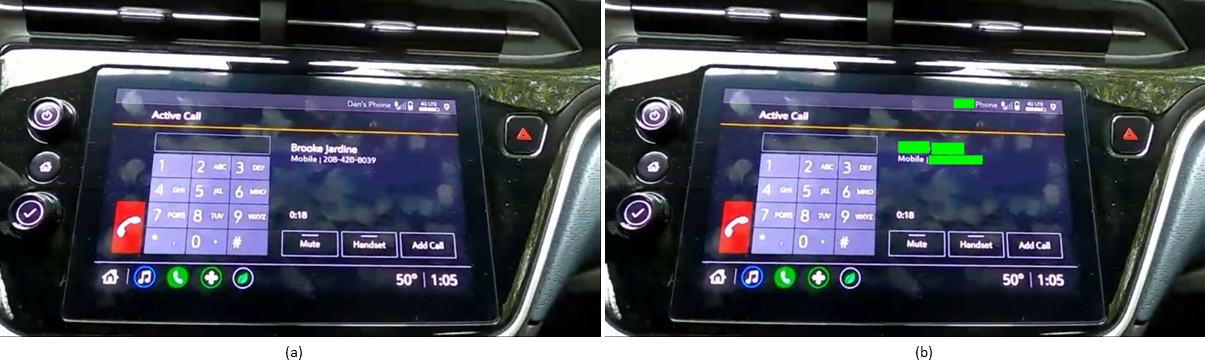}
  \caption{An example of redacting names, and phone numbers using Google Cloud Vision. (a) The image on the left is without redaction, while (b) the image on the right has names and phone numbers automatically redacted.}
  \Description{Task.}
  \label{fig:teaser}
\end{teaserfigure}


\maketitle

\section{Introduction}

% \SA{This is how we add comments. Use AM and DZ to add yours.} 

% \AM{ksdjgl}

% \DZ{Dylan's comment}

The massive improvements over the past few decades in digital camera technology, ranging from reductions in camera costs to enhancements in image quality to greater efficiency in video storage and transmission, have led to an enormous increase in the amount of available video data. This, combined with major developments in computer vision and machine learning technology, has created enormous opportunities to make life better through the collection and utilization of this video data. Potential applications here range from improved security to interactive entertainment. However, the collection and utilization of this data also entails ethical privacy concerns and the potential for unwanted intrusion into people’s lives without their permission. One way to attempt to achieve the benefits of more omnipresent video collection while mitigating the intrusion on privacy is through the automatic redaction of personally identifiable information (PII). This means automatically removing or obscuring content from video data that can be used to identify an individual while maintaining as much other video data as possible.   

A relatively new context generating a significant amount of video data is the cabins of automobiles. With the emergence of automatically responsive infotainment systems, there is the potential to improve the safety and overall driver experience by quickly integrating data from driver movement into what the infotainment system displays. However, building and improving algorithms to achieve this requires the recording and analysis of a significant amount of in-cabin video data, which drivers may be uncomfortable sharing. One way to decrease this lack of comfort is through the removal of PII from these in-cabin videos.

Within in-cabin videos, textual PII mostly appears on the infotainment system display. For example, if the infotainment system is used to call a friend or navigate to a family member’s house, the system could display PII as part of this task. The goal of this paper is to investigate how this type of PII can be automatically removed without removing other parts of the captured data. 

The data used for this investigation comes from a set of usability studies conducted by Pulse Labs AI\footnote{\url{https://pulselabs.ai}}. Pulse Labs is a company that specializes in using AI techniques to improve driver safety and experience. For these usability studies, drivers pointed cameras at their infotainment systems and recorded themselves performing common tasks like calling a friend or navigating to a business. Image (a) from Figure \ref{fig:teaser} above is an example frame from one of these videos.

Earlier research \cite{ren2018learning, gafni2019live, xu2020centerface, george2021effectiveness} on removing PII from videos has focused on anonymizing faces. In contrast, this paper focuses on redacting PII (like names, email addresses, or phone numbers) from text appearing in videos. Figure \ref{fig:teaser} shows an example of this process in action. 

Our contributions in this work can be summarized as:
\begin{itemize}
    \item We extend image redaction techniques by applying them to electronic display system interaction videos using a combination of optical character recognition (OCR) and natural language processing (NLP).
    \item We use real-world datasets to investigate the redaction of sensitive textual PII from electronic display system videos.
    \item We provide a systematic comparison of the performance of two popular OCR models, Tesseract and GCV, in the context of in-cabin video interactions.
\end{itemize}

% Extending redaction from images to videos
% First to show redaction services for a combination of different entities (names, phone, email) with validation on a real-world dataset.
% Comparison of open-source vs industrial solution

% comparing models, introducing multiple class of redaction in a single frame, testing on in-car data

\section{Proposed Approach}
\label{sect-approach}
Our approach aims to remove sensitive information from the electronic display system present within videos on a frame-by-frame basis. In this procedure, we first use an OCR model to identify text, and the box-coordinates of that text, within a frame. Next, we use NLP techniques to identify the presence of entities within the text, in our case names, phone numbers, and email addresses. Specifically, we use regular expressions\footnote{Regular expressions are a way to describe a pattern of text for which we're looking. For example, to detect the pattern x@y.z, where x, y, and z are words, we use the regular expression ``$[\backslash w\backslash.\backslash d]+\backslash @[\backslash w\backslash d]+ \backslash.[\backslash w\backslash d]+ $''.} (RegEx) to identify email addresses and phone numbers based on pattern matching, and deep learning based named-entity recognition (NER) to detect names. Finally, we apply redaction by adding green boxes at the identified coordinates of different entities. All the individual redacted frames are converted back to video by replacing their corresponding frames in the original unredacted video.

% For example, the regular expression ``$[\backslash w\backslash.\backslash d]+\backslash @[\backslash w\backslash d]+ \backslash.[\backslash w\backslash d]+ $'' detects the pattern x@y.z, where x, y, and z are words.
% $[\backslash w$\backslash$.\backslash d]+$\backslash$@[\backslash w\backslash d]+$\backslash$.[\backslash w\backslash d]+ $

\begin{figure}[h]
  \centering
  \includegraphics[width=\linewidth]{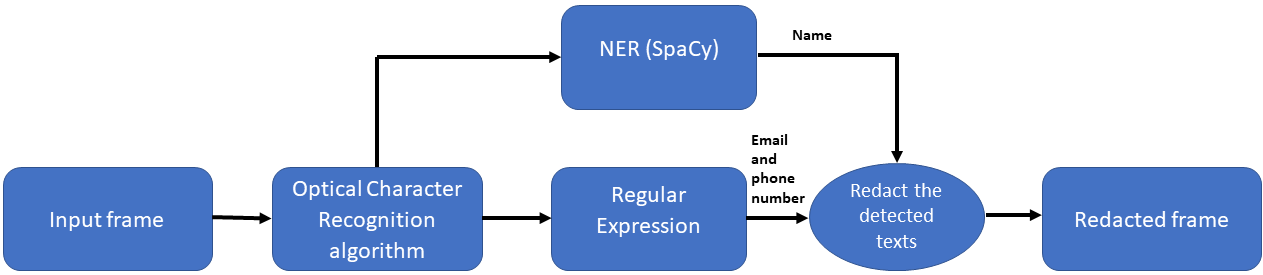}
  \caption{Pipeline for proposed redaction method.}
  \Description{Pipeline for the proposed redaction method.}
  \label{fig:fig2}
\end{figure}

In our research, we experimented with two OCR models - Tesseract\footnote{\url{https://github.com/tesseract-ocr/tesseract}} \cite{kay2007tesseract}, and Google Cloud Vision\footnote{\url{https://cloud.google.com/vision}}. The NER tool used in our approach was spaCy\footnote{\url{https://spacy.io/}} \cite{spacy2} in Python. Figure~\ref{fig:fig2} provides an overview of our complete pipeline. 

\begin{figure}[h]
\centering
\begin{subfigure}{.5\textwidth}
    \centering
    \includegraphics[width=0.98\textwidth]{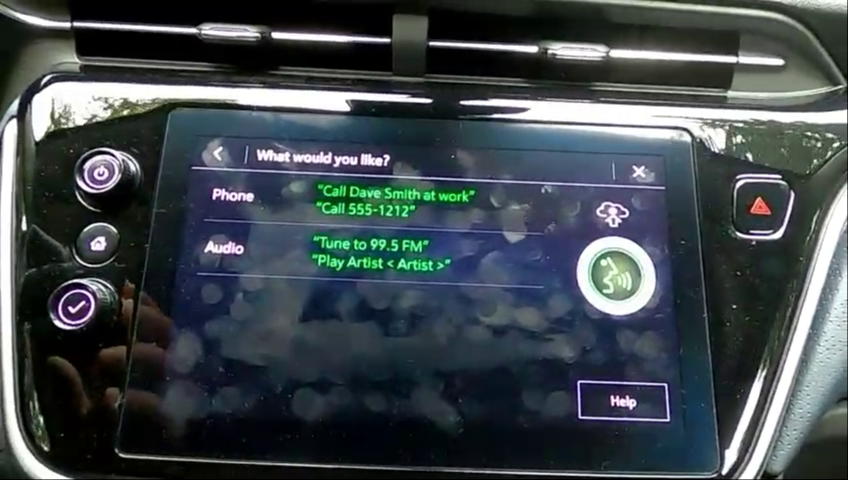}
    \caption{Input image}
    \label{fig:fig3-input}
\end{subfigure}%
\begin{subfigure}{.5\textwidth}
    \centering
    % \includegraphics[width=0.98\textwidth]{samples/pics/GT_2.png}
    % \caption{Ground truth are drawn with red bounding boxes}
    \includegraphics[width=0.98\textwidth]{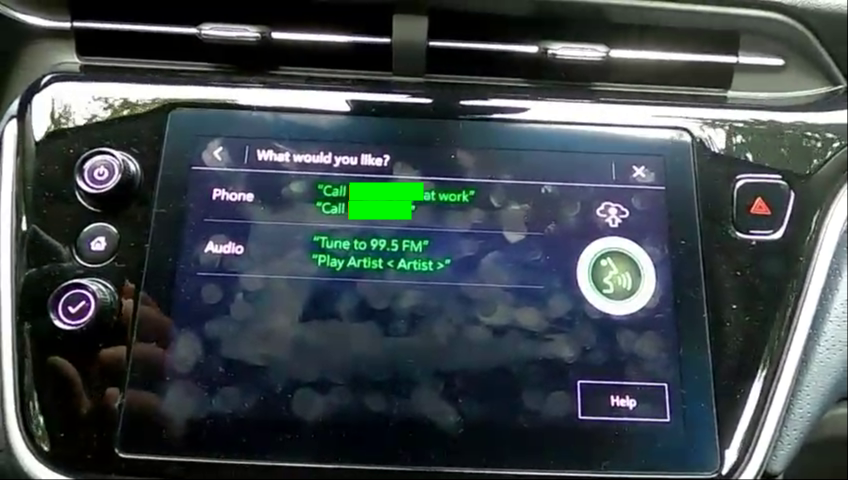}
    \caption{Ground truth labeled by human annotator}
\label{fig:fig3-gt}
\end{subfigure}
\begin{subfigure}{.5\textwidth}
    \centering
    \includegraphics[width=0.98\textwidth]{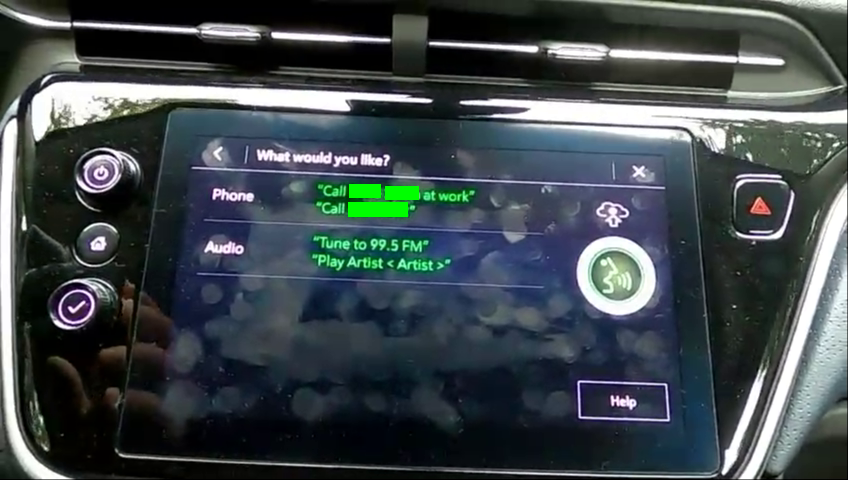}
    \caption{Output image generated by GCV}
    \label{fig:fig3-gcv}
\end{subfigure}%
\begin{subfigure}{.5\textwidth}
    \centering
    \includegraphics[width=0.98\textwidth]{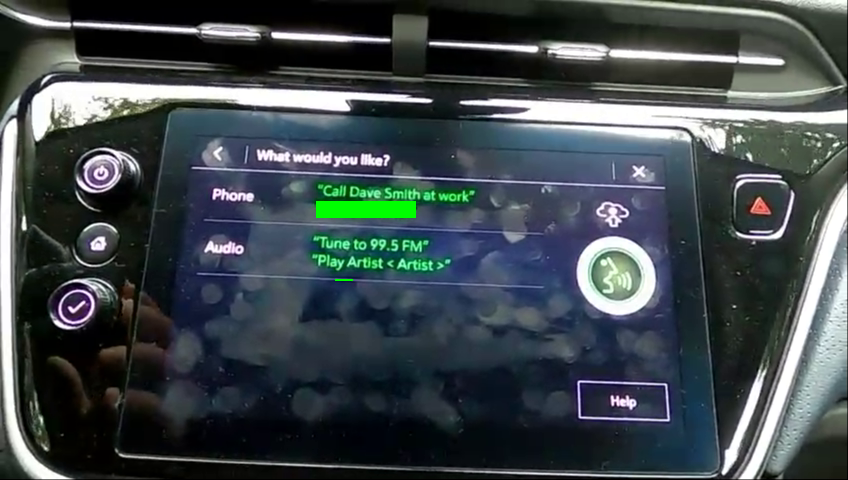}
    \caption{Output image generated by Tesseract}
    \label{fig:fig3-tesseract}
\end{subfigure}
\caption[Overview of image samples and corresponding outputs.]{Overview of image samples and corresponding outputs.}
\label{fig:fig3}
\end{figure}

An example of a frame with redaction performed using Tesseract for OCR, using GCV for OCR, and using manual identification, is provided in Figure~\ref{fig:fig3}. Here, we find that while the OCR from GCV (See Figure \ref{fig:fig3-gcv}) is able to correctly identify and anonymize both the name and phone number, the Tesseract OCR (Figure \ref{fig:fig3-tesseract}) fails to identify the name, leading to partial redaction and insufficient anonymization.

% \begin{figure}[h]
%   \centering
%   \includegraphics[width=\linewidth]{samples/pics/figure3.png}
%   \caption{Overview of image samples and corresponding outputs. (a) Input image; (b) Ground truth are drawn with red bounding boxes; (c) Output image generated by Tesseract; (d) Output image generated by GCV.}
%   \Description{Overview of image samples and corresponding outputs. (a) Input image; (b) Ground truth are drawn with red bounding boxes; (c) Output image generated by Tesseract; (d) Output image generated by GCV.}
%   \label{fig:fig3}
% \end{figure}

% We analyzed this technique using two OCR models - Tesseract\footnote{\url{https://github.com/tesseract-ocr/tesseract}}, and Google Cloud Vision \footnote{\url{https://cloud.google.com/vision}}. The NER tool used in our approach was spaCy\footnote{\url{https://spacy.io/}}. The complete pipeline is in Figure~\ref{fig:fig2} below.

%We use two OCR techniques - Tesseract\footnote{\url{https://github.com/tesseract-ocr/tesseract}} \cite{kay2007tesseract} and Google Cloud Vision\footnote{\url{https://cloud.google.com/vision}} and compare their performance in terms of accuracy. 
%Once the textual information is detected, the next task is to find target words (or entities) like email address, phone number, or name. We use regular expression~(RegEx) to detect email and phone numbers and `Spacy'\footnote{\url{https://spacy.io/}}, a named entity recognizer (NER)~\cite{spcy} for detecting names. 
% In this context, RegEx are a way to describe a pattern of word that we are looking for in the texts. For example, we constructed ``$[\backslash w$\backslash$.\backslash d]+$\backslash$@[\backslash w\backslash d]+$\backslash$.[\backslash w\backslash d]+ $'' to find email addresses in the texts and redacted when pattern matched. 

\section{Results and Discussion}
\label{sect-results}
The data for this analysis was taken from different sets of videos. Each video set featured user interactions that tended to surface textual PII of a specific type. These three types were names, phone numbers, and email addresses. From each of these sets of videos, a test dataset was created by randomly sampling 100 frames from the videos within the set. Each frame was then examined manually, and the textual entity of interest was identified and marked. The resulting datasets contained 209, 183, and 214 instances of names, phone numbers, and email addresses respectively.

We then applied our PII redaction approach to each frame twice, once using Tesseract and once using GCV. We compared the PII identifications created with the proposed approach with those created manually, and evaluated the performance of the proposed approach using metrics such as accuracy, precision, and recall. Note that for any frame it was possible to get a count of true positives, false positives, and false negatives, but not true negatives, as the number of times our approach correctly did not detect PII is not well defined. Consequently, for this analysis accuracy is defined as $TP/(TP+FP+FN)$, where $TP$, $FP$, and $FN$ indicate true positive, false positive, and false negative, respectively. Precision, $TP/(TP+FP)$, and recall, $TP/(TP+FN)$, are defined as usual. 

%We sample real-world images captured in cars to validate the performance of the proposed approach. The test dataset was created by randomly sampling 100 images from different videos for each class. The resulting dataset contains 214, 209, and 183 instances of email addresses, names, and phone numbers respectively.  We use accuracy, precision, and recall metrics to evaluate the performance. Accuracy is defined as $TP/(TP+FP+FN)$, where $TP$, $FP$, and $FN$ indicate true positive, false positive, and false negative respectively. We test the proposed methods on NVIDIA RTX 2070.

\begin{figure*}[h]
  \centering
  \includegraphics[width=\linewidth]{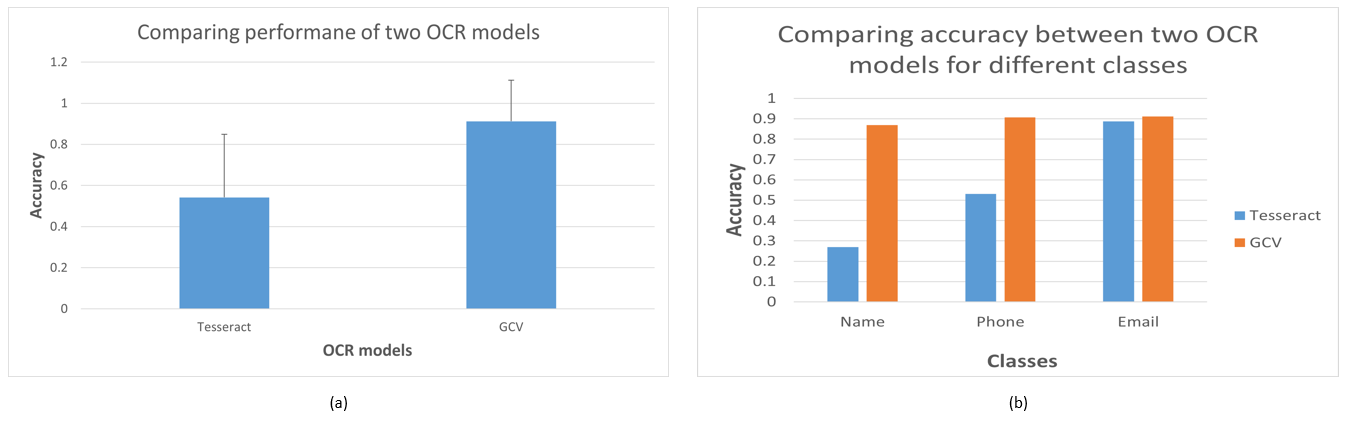}
  \caption{Accuracy analysis graphs. (a) Overall accuracy between two OCR models; (b) Class-wise performance.}
  %\Description{}
  \label{fig:fig4}
\end{figure*}

Across the three datasets, we observed that GCV (mean accuracy: $91.21\%$) obtained a significantly higher accuracy than Tesseract (mean accuracy: $54.21\%$) (Figure \ref{fig:fig4}a). For the three types of textual PII we focused on, the accuracy of Tesseract was much lower than GCV (Figure \ref{fig:fig4}b) except on email addresses, where the two models were much closer. The precision and recall of the two OCR models for every type of textual PII are given in Table \ref{tab1}. 

\begin{figure}[h]
  \centering
  \includegraphics[width=\linewidth]{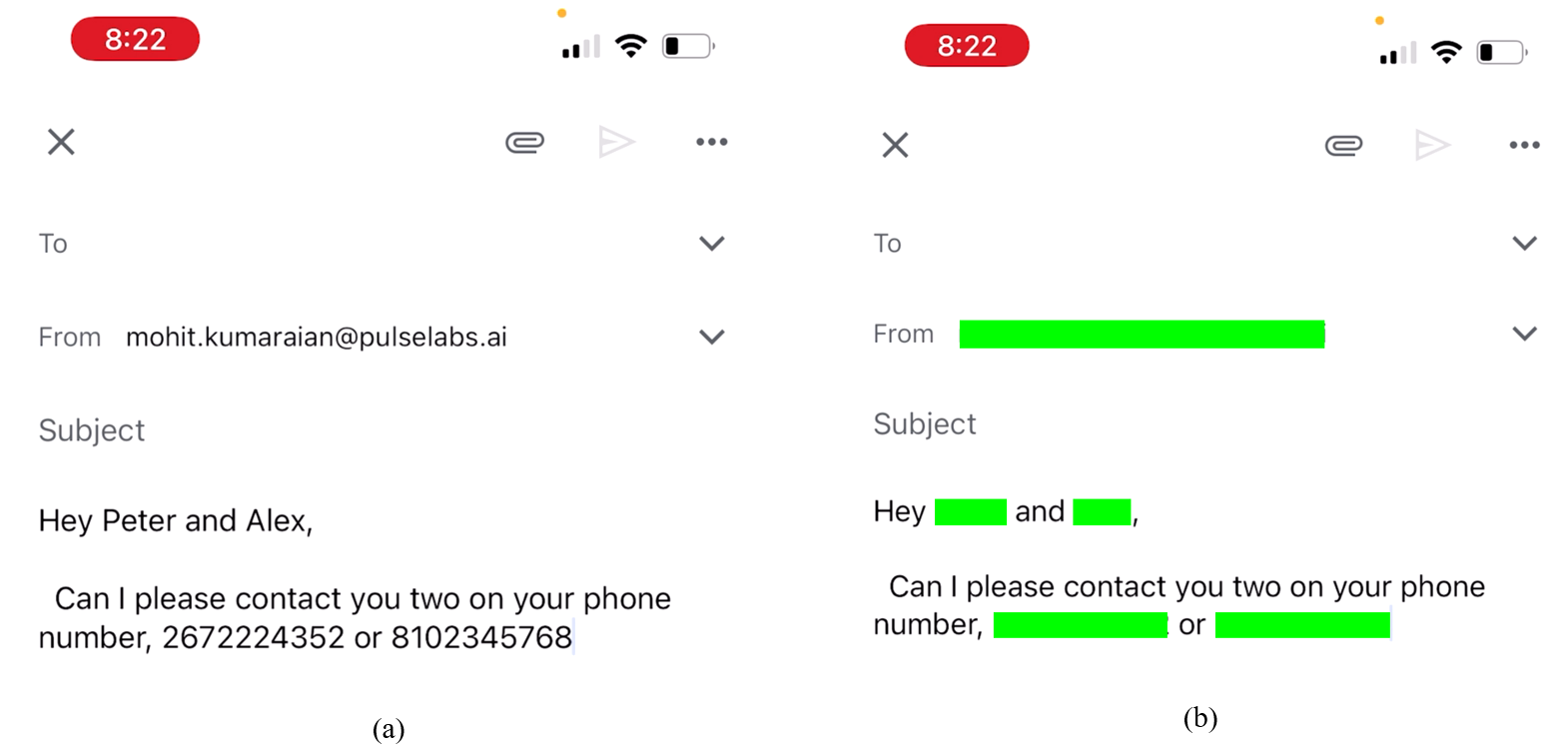}
  \caption{An example of redacting names, email addresses and phone numbers using Google Cloud Vision on mobile screen recording.}
  \Description{Pipeline for the  proposed redaction method.}
  \label{fig:mobile}
\end{figure}

Finally, we compared the average frame processing speed for each OCR model, and found that GCV took on average $1.09$ seconds to process a frame, while Tesseract on average took $1.25$ seconds. All processing was done using NVIDIA RTX 2070 processor. Note the proposed model is not limited to detecting texutal PII within automotive infotainment systems, and could be applied without changes to other types of video like mobile screen recordings (see Figure \ref{fig:mobile}).

Tesseract performed significantly worse than GCV, and also required more processing time per frame. We believe the difference in accuracy is because GCV was much better at separating punctuation within names and phone numbers (Figure~\ref{fig:teaser}). This deficiency in Tesseract significantly impeded the ability of RegEx-based pattern matching or spaCy-based NER to detect the textual PII of interest. However, this issue was not as prevalent within email addresses, which can explain why the two models were much closer for that type of textual PII. In addition, Tesseract struggled to accurately detect textual information within blurred images.

However, Tesseract is not entirely without virtues. It is open-sourced, free to use software that can be downloaded and deployed offline, while Google charges for the use of the GCV API, which can only be accessed online. If cost or offline access are priorities, Tesseract may still have some appeal.

%From our results, it is evident that the accuracy of Tesseract is significantly lower than GCV. GCV also takes lesser time in processing frames. These results give GCV advantages in real-time and real-world problems. We observe that GCV could separate punctuation (,.) from name and phone numbers (Figure~\ref{fig:teaser}) while Tesseract is not able to detect them separately. We hypothesise, in such situations, RegEx-based pattern matching (for phone number) or NER cannot produce positive results for Tesseract. Thus, accuracy is reported to be significantly lower for Tesseract for these two classes. Along with that, Tesseract fails to detect textual information accurately from blurred images, or if the text is not clear. However, Tesseract is open-sourced free to use software and can be deployed offline while Google charges for using the Google cloud vision (GCV) API which can only be accessed online.

\begin{table}
\begin{center}
\caption{Precision and Recall table for two OCR models}
\label{tab1}
\begin{tabular}{|c|*{4}{c|}}
\hline
{\textbf{Classes}} & \multicolumn{2}{c|}{\textbf{Tesseract}} &
            \multicolumn{2}{c|}{\textbf{GCV}} \\
            \cline{2-5}
            & {\textbf{Precision}} & {\textbf{Recall}} & {\textbf{Precision}} & {\textbf{Recall}} \\ 
            \hline
            \hline

Email address &	$1$ &	$0.89$ &	$1$ &	$0.91$  \\ 
\hline
Name &	$0.42$ &	$0.43$&	$0.88$ &	$0.98$  \\ 
\hline
Phone &	$0.90$	& $0.56$	& $1$	 & $0.91$ \\ 
\hline

\end{tabular}
\end{center}
\end{table}

%  \begin{table}
%   \caption{Frequency of Special Characters}
%   \label{tab:freq}
%   \begin{tabular}{ccl}
%      \toprule
%      Non-English or Math&Frequency&Comments\\
%      \midrule
%      \O & 1 in 1,000& For Swedish names\\
%      $\pi$ & 1 in 5& Common in math\\
%      \$ & 4 in 5 & Used in business\\
%      $\Psi^2_1$ & 1 in 40,000& Unexplained usage\\
%   \bottomrule
%  \end{tabular}
%  \end{table}

% \begin{equation}
%   acc = \frac{\sum_{class_{i}}C_{class}}{\sum_{class_{i}}C_{class}S_{class}}
% %   \sum_{i=0}^{\infty}x_i=\int_{0}^{\pi+2} f
% \end{equation}

%\section{Analysis}

%We use .... for our experiments and evaluation. 

%standard automatics metrics such as Accuracy, Precision and Recall are used for evaluation.

%In the following, we report results on the

\section{Conclusion}
\label{sect-conclusion}
In this paper, we proposed an approach for the automated detection of textual PII within recorded videos, based upon the use of OCR and NLP techniques. Our initial results show that we can successfully redact most of the textual entities in the recorded videos with a high accuracy. For the two OCR models, Tesseract and GCV, compared in this paper the performance of GCV with the proposed approach was significantly better. Our results are an initial step towards the automatic removal of PII from in-cabin automotive video data, which will be valuable as the need to respect driver privacy while improving the computer vision models used by automotive infotainment systems become increasingly important.

\begin{acks}
We would like to thank Abhishek Suthan, Mohit Kumaraian, Mohan Karthik and the Pulse Labs team for the fruitful discussions during the initial part of the project. 
\end{acks}

\bibliographystyle{ACM-Reference-Format}
\bibliography{sample-base}

%\appendix



\end{document}